\documentclass{article}

\usepackage{microtype}
\usepackage{graphicx}
\usepackage{subcaption}
\usepackage{booktabs}

\usepackage{hyperref}

\usepackage[accepted]{icml2024}

\usepackage{amsmath}
\usepackage{amssymb}
\usepackage{mathtools}
\usepackage{amsthm}

\usepackage[capitalize,noabbrev]{cleveref}

\theoremstyle{plain}

\theoremstyle{definition}

\theoremstyle{remark}

\icmltitlerunning{Using Neural Networks for Data Cleaning in Weather Datasets}

\usepackage{siunitx}
\usepackage{subcaption}

\usepackage{xcolor}

\newcommand{\ibtracs}{IBTrACS}
\newcommand{\unet}{U-Net}
\newcommand{\relu}{ReLU}

\graphicspath{{images/}}

\begin{document}

\twocolumn[
\icmltitle{Using Neural Networks for Data Cleaning in Weather Datasets}

\icmlsetsymbol{equal}{*}

\begin{icmlauthorlist}
\icmlauthor{Jack R. P. Hanslope}{bristol}
\icmlauthor{Laurence Aitchison}{bristol}
\end{icmlauthorlist}

\icmlaffiliation{bristol}{Department of Computer Science, University of Bristol, Bristol, UK}

\icmlcorrespondingauthor{Jack Hanslope}{jack.hanslope@bristol.ac.uk}

\icmlkeywords{Machine Learning, ICML}

\vskip 0.3in
]

\begin{NoHyper}
\printAffiliationsAndNotice{}
\end{NoHyper}

\begin{abstract}
In climate science, we often want to compare across different datasets.
Difficulties can arise in doing this due to inevitable mismatches that arise between observational and reanalysis data, or even between different reanalyses.
This misalignment can raise problems for any work that seeks to make inferences about one dataset from another.
We considered tropical cyclone location as an example task with one dataset providing atmospheric conditions (ERA5) and another providing storm tracks (\ibtracs{}).
We found that while the examples often aligned well, there were a considerable proportion (around 25\%) which were not well aligned.
We trained a neural network to map from the wind field to the storm location; in this setting misalignment in the datasets appears as ``label noise'' (i.e.\ the labelled storm location does not correspond to the underlying wind field).
We found that this neural network trained only on the often noisy labels from \ibtracs{} had a denoising effect, and performed better than the \ibtracs{} labels themselves, as measured by human preferences.
Remarkably, this even held true for training points, on which we might have expected the network to overfit to the \ibtracs{} predictions.
\end{abstract}

\section{Introduction}

In the field of climate science, there are many different datasets from many different sources.
There are often times when we would like to compare across different datasets.
For example, we might want to predict tropical cyclone tracks~\citep{richman2012adaptive,camargo2016tropical,camargo2019tropical}, flooding~\citep{mosavi2018flood,park2020prediction,motta2021mixed} or the power output of a renewable power station~\citep{foley2012current,kumar2018artificial,sharadga2020time} from atmospheric condition data.

We might, when performing these comparisons, encounter discrepancies between datasets~\citep{greatbatch2006discrepancies,li2022comparison}.
These discrepancies can arise for a number of reasons.
One reason is that some datasets are reanalyses~\citep{reanalysis_fact_sheet}.
In reanalysis the goal is often to produce an accurate, globally consistent representation of the atmospheric conditions.
However, as we cannot directly measure the atmospheric conditions, we must use computational modelling to estimate them from indirect observations.
This estimation procedure can introduce mismatches to direct observational data (e.g.\ the output of a renewable energy plant), or between different reanalyses.
These discrepancies can make it is difficult to draw connections between variables in different datasets, which causes significant problems if the two variables of interest in our scientific analysis are only available in different datasets.

We studied how to mitigate these issues of mismatch between different datasets.
As an example, we studied tropical cyclones.
We had two reanalyses, one providing the location of tropical cyclones \citep[\ibtracs{};][]{ibtracks_paper,ibtracks_dataset} and one providing the atmospheric conditions \citep[ERA5;][]{era5_dataset,era5_paper}.
Initially, we were interested in predicting the future location of a tropical cyclone, when provided with the wind field in the present.
However, preliminary analyses indicated that while the track data and the wind field often aligned well, there were many occasions where this was not the case, even when comparing the wind field and tropical cyclone location at the same time.
There were a number of ways in which this misalignment appeared.
In some instances, ERA5 would show very obvious signs of a tropical cyclone in one location while \ibtracs{} would indicate a tropical cyclone in a different location.
Sometimes, there would appear to be more storms within the wind field than the number indicated by the storm location data.
If we consider the ERA5 wind field as the input and the \ibtracs{} storm location as the output, then we can consider this to be an issue of having noisy labels.

Given that there is this misalignment between the \ibtracs{} storm locations and the ERA5 wind field, it may seem that the only solution would be to hand-label the storm locations, based on the ERA5 wind fields.
We considered however, whether we could leverage the fact that, in around 75\% of instances, there was a very close match between the two datasets.
In other words, instances where the \ibtracs{} tropical cyclone location would match the ERA5 wind field.
We therefore took a naive approach: training a neural network to predict the \ibtracs{} storm location from the ERA5 wind field using a \unet{} architecture~\citep{unet}.
Surprisingly, despite being trained on often noisy/mismatched data, we found that the neural network almost always matched human judgements of storm location based on the ERA5 wind field.
Indeed, even in instances where the two datasets were misaligned, the neural network usually gave a much better predication of the location than the tropical cyclone location dataset.

\section{Related Work}
There has been much research in the field of noisy label machine learning~\citep{natarajan2013learning,han_2019_iccv,song2022learning}.
However, none of this work has studied whether neural networks are able to address the issues raised by the noisy labels that are inevitably observed in weather datasets.

\section{Methodology}

\subsection{Data Description}

\subsubsection{ERA5}
For this work, we need a dataset providing the atmospheric conditions.
Such data includes variables like pressure, temperature, wind speed and direction, and atmospheric pressure.
We use ERA5 to provide this data~\citep{era5_dataset,era5_paper}.
ERA5 is provided by the Copernicus Climate Change Service (C3S) at the European Centre for Medium-Range Weather Forecasts (ECMWF).

It is fairly obvious that it is impossible to obtain measurements of every variable we might be interested in at every location (latitude, longitude, altitude) within the Earth's atmosphere.
As such, ERA5 is a reanalysis dataset.
A reanalysis first takes a weather forecasting model and uses it to create a prediction for a certain weather variable (for example, temperature) based on the value of that variable at some earlier time.
It then corrects this prediction using known, but indirect, observations; this step is called data assimilation~\citep{dee2011era}.
A reanalysis provides a globally consistent picture of the atmospheric conditions at any given time, but it is ultimately an estimate based on indirect measurements and modelling.

The ERA5 reanalysis provides many different variables which are either provided at a single level or at different pressure levels.
The different pressure levels are at 37 pressure levels from \SI{1000}{\hecto\pascal} to  \SI{1}{\hecto\pascal} (hectopascals).
Pressure varies inversely with altitude from sea level, that is, as altitude increases, pressure decreases.
Most of the ERA5 variables are provided hourly, at a \SI{31}{\kilo\metre} resolution.

Using all of the variables from the dataset would require an enormous amount of storage, so instead we use only a selection.
We use wind data because this is, intuitively, the most indicative of a hurricane.
We have $u$ (eastward) and $v$ (northward) components of the wind.
We use the wind at \SI{850}{\hecto\pascal} which corresponds with an altitude of approximately \SI{1500}{\metre}.

It is worth noting that the aim of this work is not to produce the most accurate storm location model possible.
If that were the goal, then we would almost certainly be able to achieve a higher accuracy if we used more variables.
The goal of this work is to explore whether we are able to use a neural network to mitigate the mismatch between different reanalysis datasets.

\subsubsection{\ibtracs{}}
In addition to a dataset providing us with the atmospheric conditions, we also require a dataset providing us with locations of tropical cyclones.
The dataset we selected is the International Best Track Archive for Climate Stewardship \citep[\ibtracs{};][]{ibtracks_paper, ibtracks_dataset}.

This dataset combines best track datasets from many different national agencies around the world.
An agency's best track dataset provides the best location estimate for each point in the storm's lifetime, as well as storm intensity at each of those points.

The \ibtracs{} dataset provides many variables but the only information we use from this dataset is the latitude and longitude of the tropical cyclone centre.
This information is usually provided at 3 hour intervals, with some additional timestamps (for example, when a tropical cyclone makes landfall).

\subsection{Neural Network Setup}
The neural network takes as input the ERA5 $u$ and $v$ channels of \SI{850}{\hecto\pascal} wind speeds at one timestamp.
The network output is an estimate of the location of the tropical cyclone, and we train the network to match the locations given by \ibtracs{}.
While we could return this estimate as a continuous latitude and longitude value (i.e.\ using regression), we instead ask the neural network to classify the grid box that the tropical cyclone is located within, as this allows the network to naturally return an estimate of the probability that the tropical cyclone is within a given grid box.

We restrict the data to only the North Indian basin and only instances where there is exactly one storm in the basin.
The basin is constrained by the equator to the south (all latitudes are greater than 0) and longitudes are between 30$^\circ$ and 100$^\circ$ east.
Only tropical storms since 1980 were used, since prior to then, identification of storms was less reliable.
Since then, satellites have been used to locate storm centres.
We discard any time steps that do not fall on a whole number of 3 hours.
In order to reduce the storage requirements of the data, we coarsen the data to the nearest degree (coarsening by a factor of about 4).

\subsection{Neural Network Architecture}
The model we trained for storm location is a \unet{}~\citep{unet}.
These typically perform well on tasks with image-like outputs; our network outputs an ``image'' giving the probability of the tropical cyclone being within each grid box.

\unet{}s have an encoder-decoder structure and the encoder half of our network has 4 downsampling layers.
Each has two $3\times3$ convolutions followed by \relu{}s and $2\times2$ max pooling.
Note that the final layer has no max pooling.
The decoder layers consist of interpolation upsampling concatenated to the output of the corresponding layer in the encoder (skip connections) and two $3\times3$ convolutions, each followed by a \relu{}.
The number of filters in the encoder layers are 64, 128, 256 and 512; similarly, the decoders have 512, 256, 128 and 64 filters.

We use the \unet{} implementation provided in the GitHub repository attached to \citet{3d_unet_github_paper}.

\subsection{Training Setup}
The batch size was 32 and we trained for 100 epochs.
We used Adam~\citep{adam_optimizer} with a learning rate of $10^{-3}$ on a cross entropy loss.
We split the data into training, validation and test with 15\% each for the test and validation sets and 70\% for the training set.
At each learning step, the network would output a logit for each of the $32 \times 56 = 1792$ grid boxes, describing the probability of the tropical cyclone being within that grid box.
We calculate the cross entropy loss by comparing this to the target cell from \ibtracs{}.

Post-hoc, we apply temperature scaling~\citep{guo2017calibration}.
This method is intended to bring the confidences of the network closer to the true probabilities.
It scales all of the network's logits by a single number, the temperature, which is tuned on the validation set.
The relative order of the class scores is maintained.

\section{Results}

Now, we compare the original \ibtracs{} storm locations to the locations provided by our trained network.
Overall, we find that when they suggest different locations, usually the neural network gives a better location.

\subsection{Qualitative Results}

\begin{figure*}[t]
  \centering
    \begin{subfigure}{0.49\textwidth}
      \includegraphics[width=\textwidth]{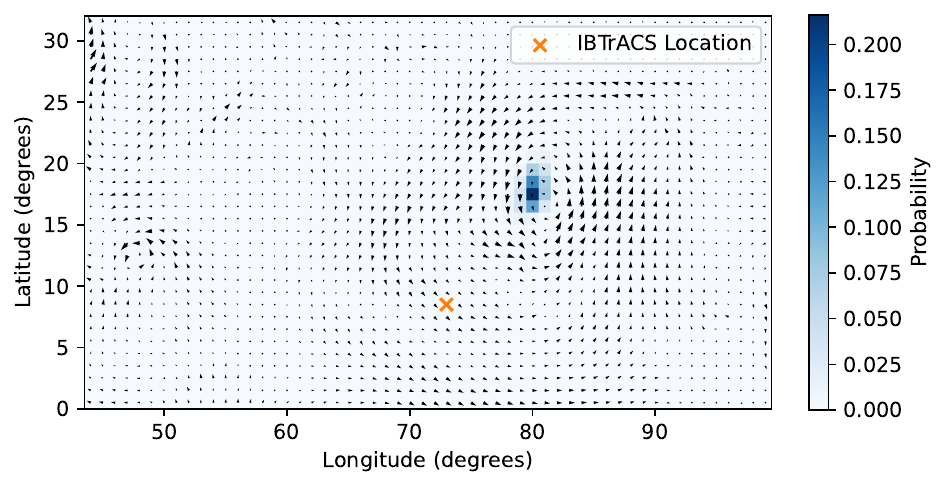}
    \end{subfigure}
    \begin{subfigure}{0.49\textwidth}
      \includegraphics[width=\textwidth]{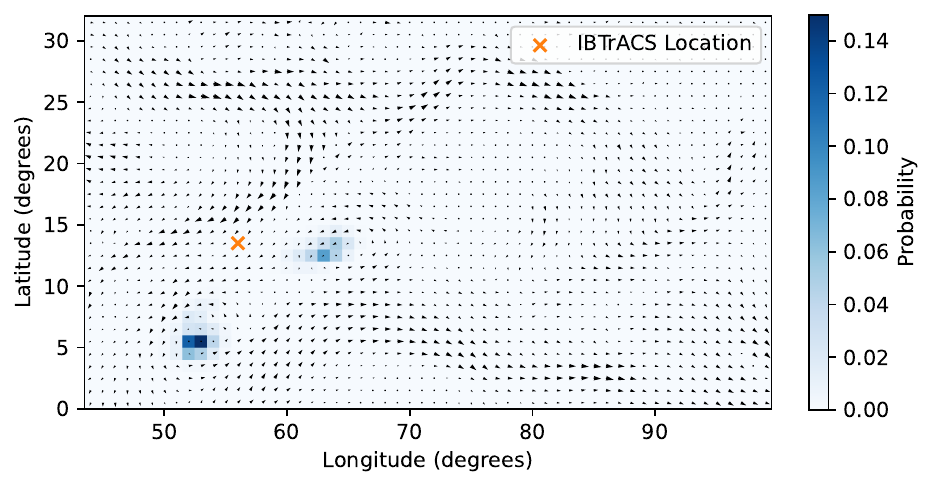}
    \end{subfigure}
    \begin{subfigure}{0.49\textwidth}
      \includegraphics[width=\textwidth]{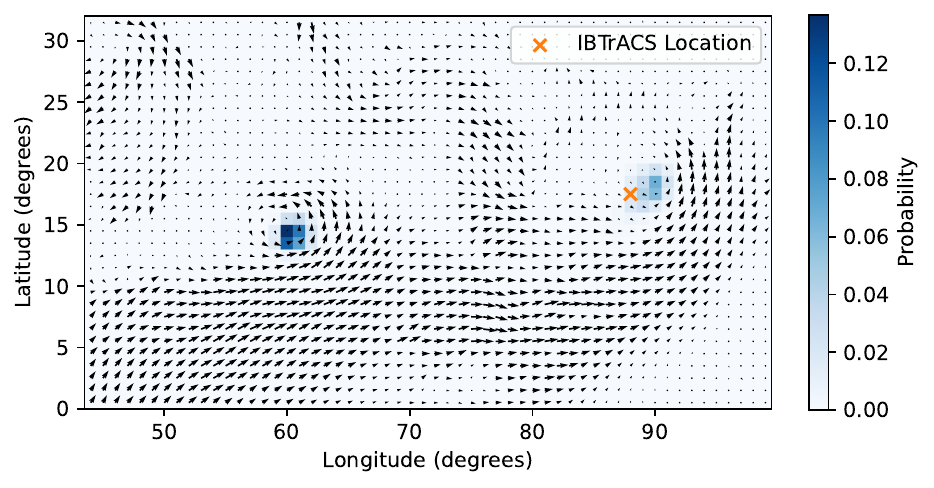}
    \end{subfigure}
    \begin{subfigure}{0.49\textwidth}
      \includegraphics[width=\textwidth]{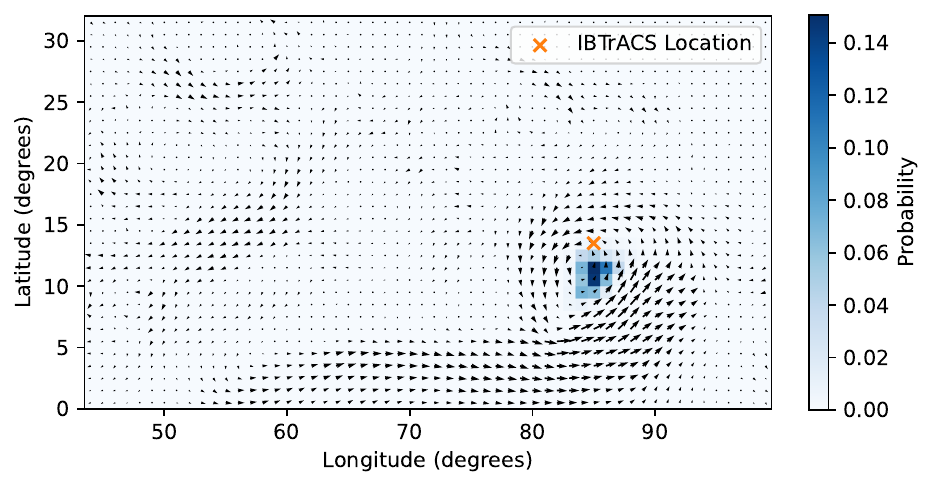}
    \end{subfigure}
    \caption{Wind fields for 4 hand chosen storms\@. \ibtracs{} location of storm centre marked by orange cross and probability predicted by the \unet{} shown by the colourmap. Top left 1980-10-19 03:00; top right 1986-11-11 09:00; bottom left 1987-06-02 21:00; bottom right 1996-11-28 18:00.}\label{figure:wind_fields} 
\end{figure*}

\cref{figure:wind_fields} shows four wind fields
The wind direction and magnitude is show by the direction and size of the arrows on each point; this data comes from ERA5.
The location of the storm centre, provided by \ibtracs{} is shown by the orange cross.
The probabilistic predictions of the model are shown as a colourmap, ranging from white (low probability) to dark blue (high probability).

The particular storms were chosen by hand to show a mismatch between \ibtracs{} and ERA5.
In most of these instances, we see that the model will usually put most of its confidence in the middle of whichever cyclone has the strongest wind.
Intuitively, this is where we, as humans, would expect the storm to be.
In some of these wind fields, the ground truth label does not appear to be in the middle of a cyclone, and in some cases does not appear to be in a cyclone at all.

\subsection{Quantitative Results}
\begin{figure}[ht!]
\centering
\includegraphics[width=0.4\textwidth]{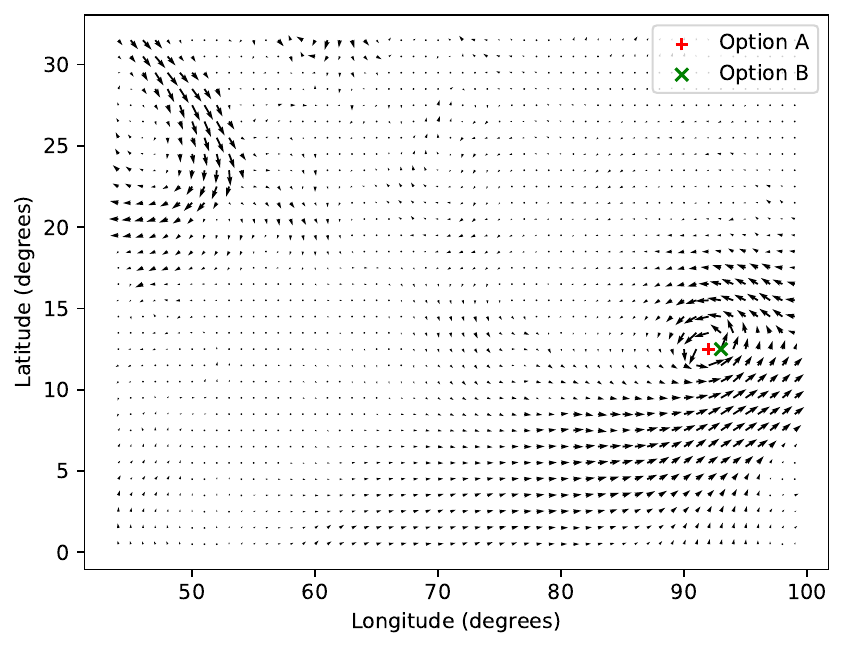}
\caption{Example of a wind field shown to the author for comparison.}\label{figure:compare}
\end{figure}

Of course, it is not possible to reach any definitive conclusions based on just those examples.
At the same time, it is difficult to know how to quantitatively compare the neural network outputs with those of \ibtracs{} in the absence of any ``ground truth'' data.
As such, to determine whether the locations provided by the \unet{} are indeed better than the locations provided by \ibtracs{} we devised a test where one of the authors was shown a wind field with two storm locations.
One of the locations was from the \unet{} and the other from \ibtracs{} but it was not indicated which was which (i.e.\ the author was blinded).
The author was asked which, if either, they preferred.
An example is shown in \cref{figure:compare}.

200 wind fields were shown from each of the train and test set, with the storms being chosen randomly (with a seed of 0).
Note that there are many cases where the locations co-coincided, so no preference could be made.
The results are summarised in \cref{tab:test_results}, and clearly show an improvement for the \unet{}.
In particular, the human preferred the \unet{} 46 out of 59 times where there was a preference for the test set, and 49 of 64 times for the training set.

Using the 59 and 64 instances where there was a preference, we conducted one-tailed binomial tests.
The null hypothesis was that \textit{when} there was a preference of one location, both locations are equally likely.
The alternative hypothesis was that when there was a preference of one location, the \unet{} was more likely to be preferred.
The p-values are \num{9.6e-6} and \num{1.2e-5} which both strongly suggests that when there is a noticeable difference between the two locations, the \unet{} is providing a better location.
Note that it is remarkable that we see improvements even on the training set, as we might have expected the neural network to overfit to the training set, and therefore to simply give the same location as the original \ibtracs{} location that the network was trained on.

\begin{table}[t]
\caption{Preferences of \unet{} location, \ibtracs{} location and no preference for test set and train set. 200 images were shown for each set. The $p$-value is calculated as a one tailed binomial test using only the instances where one or other location was preferred.}\label{tab:test_results}
\vskip 0.15in
\begin{center}
\begin{small}
\begin{sc}
\begin{tabular}{lcccr}
\toprule
Preference & Test Set     & Train Set    \\
\midrule
\unet{}    &  46          &  49          \\
\ibtracs{} &  13          &  15          \\
Neither    & 141          & 136          \\
\midrule
Total      & 200          & 200          \\
\midrule
$p$-value  & \num{9.6e-6} & \num{1.2e-5} \\
\bottomrule
\end{tabular}
\end{sc}
\end{small}
\end{center}
\vskip -0.1in
\end{table}

\section{Conclusion}
In summary, we've shown that there are sometimes discrepancies between climate science datasets, and these discrepancies can often make it difficult to draw conclusions about a target variable in one dataset from data in the other dataset.
We have shown that it is sometimes possible to mitigate these mismatches by exploiting the denoising properties inherent in neural networks.

\bibliography{paper}
\bibliographystyle{icml2024}

\end{document}